\documentclass[11pt,a4paper]{article}

\usepackage[utf8]{inputenc}
\usepackage[T1]{fontenc}
\usepackage{hyperref}
\usepackage{url}
\usepackage{booktabs}       % professional-quality tables
\usepackage{amsfonts}       % blackboard math symbols
\usepackage{nicefrac}       % compact symbols for 1/2, etc.
\usepackage{microtype}      % microtypography
\usepackage{xcolor}
\usepackage{graphicx}       % for figures
\usepackage{float}          % better figure placement
\usepackage{subcaption}
\usepackage{amsmath}        % advanced math
\usepackage{amssymb}

\usepackage[numbers]{natbib}
\usepackage[margin=1in]{geometry}

\title{\textbf{VajraV1 - The most accurate Real Time Object Detector of the YOLO family}}

\author{
  \textbf{Naman Makkar} \\
  \textbf{Vayuvahana Technologies Private Limited} \\
  \texttt{namansingh2803@gmail.com} \\
  \href{https://github.com/NamanMakkar/VayuAI}{https://github.com/NamanMakkar/VayuAI} \\
  \thanks{This paper is licensed under CC BY 4.0. VajraV1 code and models are licensed under AGPLv3.0. Commercial enterprise licensing is available from Vayuvahana Technologies Private Limited.} \\
  \texttt{-- Technical Report --}
}

\date{\today}

\begin{document}

\maketitle

\begin{figure}[H]
    \centering
    \begin{subfigure}[b]{0.6\textwidth}
        \includegraphics[width=\textwidth]{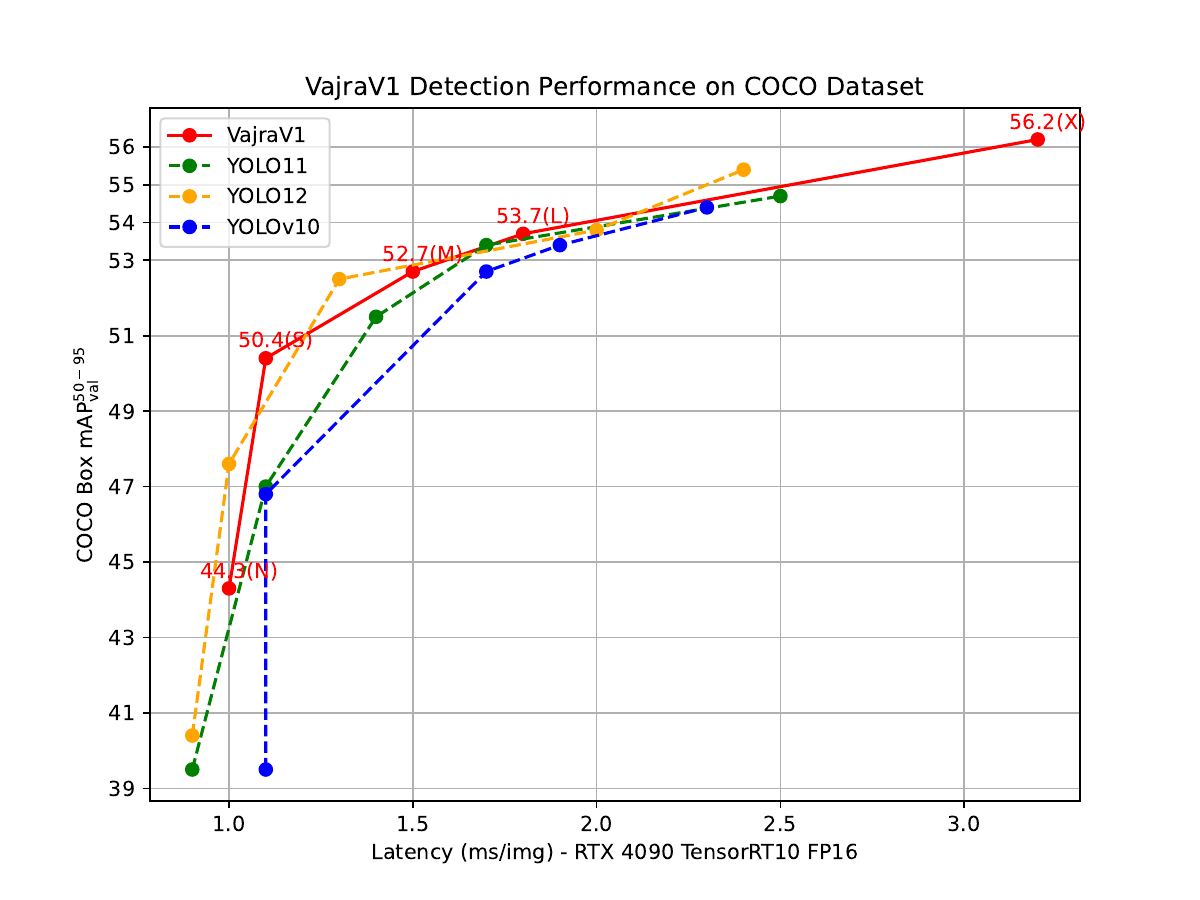}
        \caption{Detection COCO Validation AP vs Latency}
        \label{fig:det_plot}
    \end{subfigure}
    \hfill
    \begin{subfigure}[b]{0.496\textwidth}
        \includegraphics[width=\textwidth]{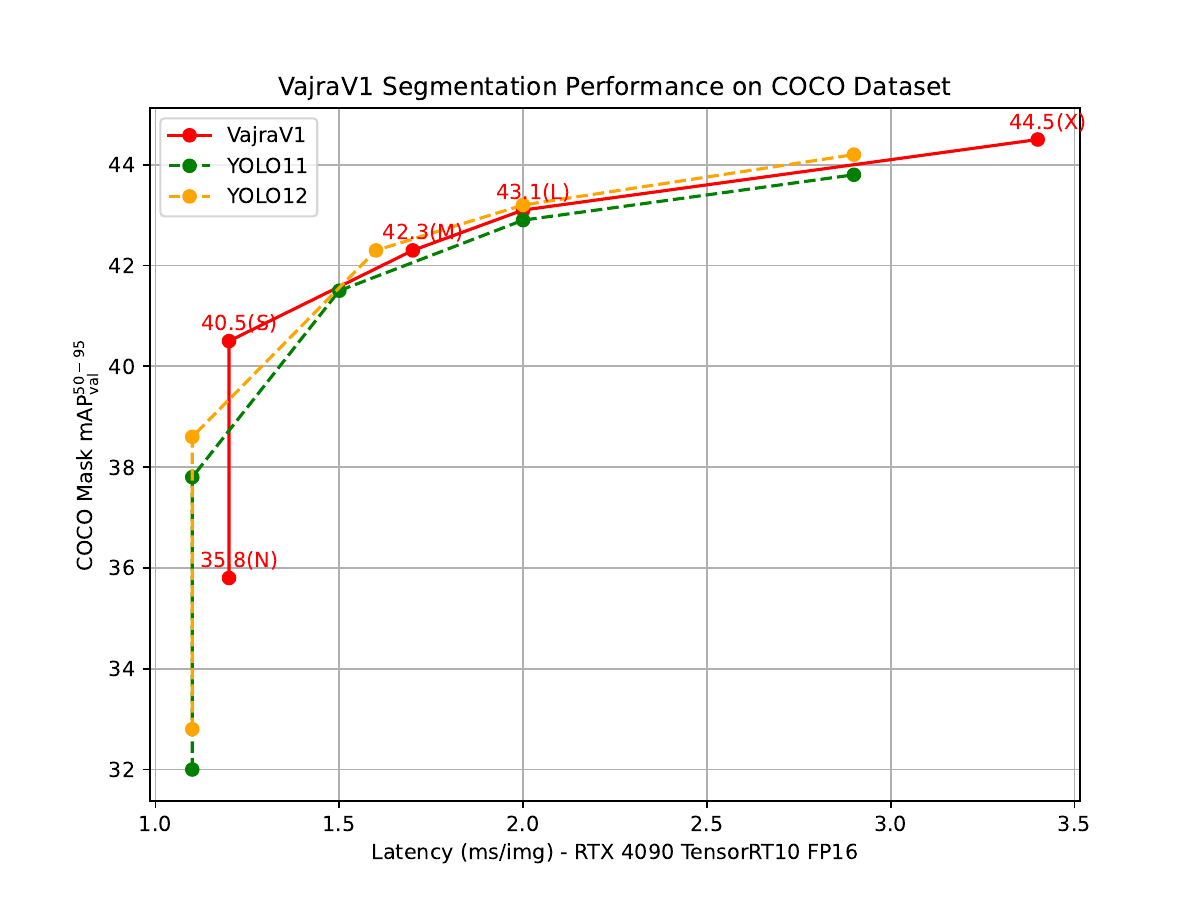}
        \caption{Segmentation COCO Validation AP vs Latency}
        \label{fig:seg_plot}
    \end{subfigure}
    \hfill
    \begin{subfigure}[b]{0.496\textwidth}
        \includegraphics[width=\textwidth]{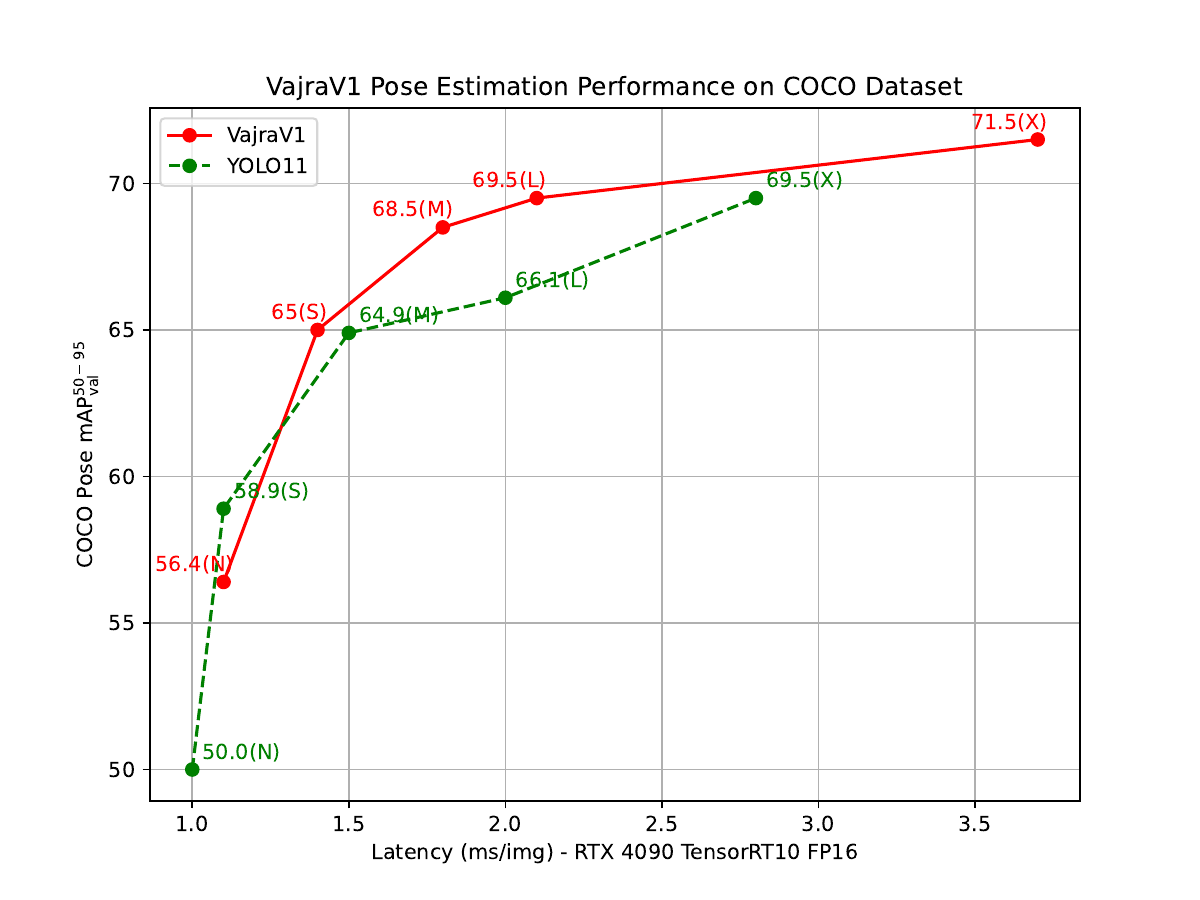}
        \caption{Pose COCO Validation AP vs Latency}
        \label{fig:pose_plot}
    \end{subfigure}
    \caption{Comparison with YOLOv11 and YOLOv12}
    \label{fig:vajra_v1_comparison}
\end{figure}

\begin{abstract}

  We have seen many advances in the field of real-time object detection in the last few years. From 2024-2025, we have witnessed the release of YOLOv10, YOLO11, YOLOv12 and YOLOv13. It is my pleasure to present to you the technical report of the VajraV1 model architecture which features enhancements over the YOLO model architectures. The VajraV1 combines the best features of the existing YOLO models to surpass all real-time object detectors in accuracy with competitive inference speed. \\ 
  
  \textbf{VajraV1-Nano} achieves \textbf{44.3\% mAP} on the COCO validation set outperforming YOLOv12-N by \textbf{3.7\%}, the YOLOv13-N by \textbf{2.7\%} with a comparable inference speed. \\

  \textbf{VajraV1-Small} achieves \textbf{50.4\% mAP} on the COCO validation set beating YOLOv12-S and YOLOv13-S by \textbf{2.4\%}. \\

  \textbf{VajraV1-Medium} achieves \textbf{52.7\% mAP} on the COCO validation set outperforming the YOLOv12-M by \textbf{0.2\%}. \\

  \textbf{VajraV1-Large} achieves \textbf{53.7\% mAP} on the COCO validation set outperforming the YOLOv13-L by \textbf{0.3\%}. \\

  \textbf{VajraV1-Xlarge} achieves \textbf{56.2\% mAP} on the COCO validation set outperforming all real-time object detectors.
  
\end{abstract}

\section{Introduction}
In the field of real-time object detection, the YOLO \cite{8100173, 7780460, redmon2018yolov3incrementalimprovement, yolov5, yolov8_ultralytics, yolo11_ultralytics, wang2024yolov9learningwantlearn, bochkovskiy2020yolov4optimalspeedaccuracy, li2023yolov6v30fullscalereloading, wang2022yolov7trainablebagoffreebiessets, lei2025yolov13realtimeobjectdetection, wang2024yolov10realtimeendtoendobject, xu2022ppyoloeevolvedversionyolo, wang2023goldyoloefficientobjectdetector, xu2023damoyoloreportrealtime} family of single-stage object detectors has established itself as the industry standard due to its optimal tradeoff between accuracy and inference speed. Over the years, we have witnessed iterative improvements from the first YOLO model to YOLOv12 \cite{tian2025yolov12attentioncentricrealtimeobject} and YOLOv13 \cite{lei2025yolov13realtimeobjectdetection}. Among these, the most critical improvements have been in the realm of network architecture design. Various primary computational blocks have been researched over the year for the purpose of optimizing the feature extraction ability of the backbone and the feature fusion ability of the neck of the object detector while minimizing the latency. These include DarkNet \cite{7780460, 8100173, redmon2018yolov3incrementalimprovement}, CSPNet \cite{wang2019cspnetnewbackboneenhance}, ELAN \cite{wang2022yolov7trainablebagoffreebiessets} and GELAN \cite{wang2024yolov9learningwantlearn}. Although most network architecture design improvements involved improving the CNN dominated primary computational block, starting with the YOLOv10 \cite{wang2024yolov10realtimeendtoendobject} we have witnessed the integration of transformer modules in the backbone of the YOLO models. Attempts have been made to integrate self-attention into the YOLO backbones in a parameter efficient manner. However, due to the quadratic computational complexity of the attention mechanism, the integration of self-attention is achieved by reducing the computational cost in other parts of the model. For example, in the YOLOv10 the downsampling 3x3 convolutional layers are replaced with depthwise-separable convolutions with the help of spatial-channel decoupled downsampling (a pointwise convolution followed by a depthwise convolution), while in the YOLOv11 \cite{yolo11_ultralytics} this is done by optimizing the width and depth scaling in the backbone and the neck of the model. This presents a unique challenge when it comes to optimizing the model architecture design, as the designer needs to find the optimal trade-off in order to integrate self-attention mechanisms in a parameter-efficient manner while not compromising the accuracy of the model.\\ 

This paper aims to address this challenge and attempts to come up with an optimal trade-off to improve the accuracy of the YOLO model family by modifying the primary computational blocks, integrating parameter efficient computational blocks and efficiently integrating self-attention into the backbone of the object detector in order to improve the performance of the object detector.\\

In order to achieve this, we take advantage of the following architectural improvements. \textbf{First}, we utilize a primary computational block (\textbf{VajraV1MerudandaX} inspired by YOLOv9's \cite{wang2024yolov9learningwantlearn} \textbf{RepNCSPELAN4}) that accommodates RepVGGBlock \cite{ding2021repvggmakingvggstyleconvnets} modules and increase the width of all 3x3 convolutions in the primary computational block to twice that of the 3x3 convolutions used in the primary computational blocks of the YOLOv11 \cite{yolo11_ultralytics} and YOLOv12 \cite{tian2025yolov12attentioncentricrealtimeobject}. \textbf{Second}, in order to make up for the increase in FLOPs (due to an increase in width) we make use of parameter efficient computational blocks (\textbf{VajraV1MerudandaBhag15} inspired by the \textbf{C2fCIB} module of the YOLOv10 \cite{wang2024yolov10realtimeendtoendobject}) taking advantage of the low-intrinsic rank in the deeper stages of the architecture and following the rank-guided block design philosophy of the YOLOv10 \cite{wang2024yolov10realtimeendtoendobject}. \textbf{Third}, once again taking advantage of the low intrinsic rank of the YOLO models we make use of FLOP efficient downsample convolutions (\textbf{ADown} first introduced in the YOLOv9 \cite{wang2024yolov9learningwantlearn}) which is used in the S5 stage of the backbone and the P5 stage of the neck of the \textbf{VajraV1-Medium} and \textbf{VajraV1-Large} models and used throughout the \textbf{VajraV1-Xlarge} model. \textbf{Fourth}, we accommodate Transformer modules in an ELAN block (\textbf{VajraV1AttentionBhag6}) identical to the \textbf{C2PSA} module used in the YOLOv11 \cite{yolo11_ultralytics} which improves the global representation learning ability of the models. Based on these approaches we come up with a family of real-time detectors with multiple model scales, i.e, VajraV1-Nano, Small, Medium, Large, Xlarge. \textbf{The results of our experiments demonstrate that VajraV1 achieves state-of-the-art performance on the MS COCO \cite{lin2015coco} Detection, Segmentation and Pose Estimation datasets.}

\section{Related Work}
This paper focuses on architectural improvements in real-time object detectors, specifically the YOLO model architectures. In recent iterations of the YOLO models, the YOLOv9 \cite{wang2024yolov9learningwantlearn} introduced the GELAN (Generalized Efficient Layer Aggregation Network) as the primary computational block of the model architecture, achieving a significant boost in performance over its predecessors. The YOLOv10 \cite{wang2024yolov10realtimeendtoendobject} introduced the Compact Inverted Block (CIB), the Positional Self-Attention (PSA) Block and a lightweight classification head consisting of depthwise separable convolutions. Additionally, YOLOv10 utilized NMS-free training with dual assignments for efficiency gains. YOLOv11 on the other hand integrated the GELAN architecture (C3K2) with the use of C3 blocks (first introduced in YOLOv5 \cite{yolov5}) as the inner blocks (C3K) and optimized the width and depth scaling to accommodate the PSA blocks of the YOLOv10 in a parameter and FLOP-efficient manner.

\subsection{YOLOv9}
YOLOv9 \cite{wang2024yolov9learningwantlearn} brought the following architectural improvements to the YOLO model family:

\begin{enumerate}
    \item \textbf{Use of FLOP efficient downsample convolutions:} The YOLOv9 models make use of the ADown convolution shown in Figure \ref{fig:adown}. This is a FLOP efficient downsample convolution that is used to replace 3x3 downsample convolutions in different stages of the backbone and neck of the model.
    \item \textbf{Generalized Efficient Layer Aggregation Network:} The GELAN module is a combination of a CSP module \cite{wang2019cspnetnewbackboneenhance} and an ELAN module \cite{wang2022yolov7trainablebagoffreebiessets}. It takes the architecture of the ELAN module (which originally only used stacking of convolutional layers) and converts it into a generalized ELAN that can also integrate various computational blocks.
    
\end{enumerate}
\subsection{YOLOv10}
YOLOv10 \cite{wang2024yolov10realtimeendtoendobject} made use of the following to improve the performance of the YOLO models:

\begin{enumerate}
    \item \textbf{Compact Inverted Block (CIB):} CIB is a lightweight computational block that is used in the YOLOv10 models. It is an inverted residual block with 3x Depthwise Convolutions (3x3) and 2 pointwise convolutions for channel mixing. For the Nano and Small models, the second 3x3 Depthwise convolution is replaced with a 7x7 Depthwise convolution.
    \item \textbf{Partial Self-Attention (PSA) Block:} This is a GELAN \cite{wang2024yolov9learningwantlearn} block that integrates a transformer module (multi head self-attention layer and a feed-forward network / MLP). This improves the global representation learning capability of the model.
    \item \textbf{Light Classification Head:} The classification branch of YOLOv10's detection head integrates lightweight depthwise-separable convolutions. This significantly reduces the overall FLOPs and latency of the model. This design choice was continued in the YOLOv11 \cite{yolo11_ultralytics}, YOLOv12 \cite{tian2025yolov12attentioncentricrealtimeobject}, YOLOv13 \cite{lei2025yolov13realtimeobjectdetection} and the VajraV1 models. 
\end{enumerate}

\subsection{YOLOv11}
YOLOv11 \cite{yolo11_ultralytics} introduced the following improvements:

\begin{enumerate}
    \item \textbf{Use of C3K2, An efficient primary computational block:} C3K2 is a GELAN inspired by the YOLOv9's \cite{wang2024yolov9learningwantlearn} RepNCSPELAN4 block. It uses an inner block called C3K similar to how RepNCSPELAN4 uses the RepNCSP inner block. However, C3K uses traditional residual bottlenecks instead of RepVGG convolutions. Each C3K module accommodates two residual bottlenecks each of which uses two 3x3 convolutions. For nano, small and medium models of the YOLOv11 each C3K2 block uses one C3K module, using a total of four 3x3 convolutions per C3K2 block, but for large and xlarge models, two C3K blocks are used giving eight 3x3 convolutions per C3K2 block. This gives the C3K2, two more 3x3 convolutions than the RepNCSPELAN4 block for the L and X models resulting in a larger receptive field than the previous YOLO models.
    
    \item \textbf{Use of C2PSA, An Efficient Layer Aggregation Network for integrating Transformer Modules:} C2PSA is a GELAN \cite{wang2024yolov9learningwantlearn} that integrates the PSA modules used in the YOLOv10. Unlike the YOLOv10 \cite{wang2024yolov10realtimeendtoendobject} which only uses one PSA module in each of its models, the YOLOv11 accommodates two PSA modules in the L and X models, giving the YOLOv11 better global representation learning ability than all of the previous YOLO models.

\end{enumerate}

\subsection{YOLOv12}
YOLOv12 \cite{tian2025yolov12attentioncentricrealtimeobject} introduced the following improvements to outperform the other YOLO models:

\begin{enumerate}
    \item \textbf{Area Attention:} This is an efficient implementation of self-attention, where a featuremap with resolution $(H, W)$ is divided into $l$ segments of size $(\frac{H}{l}, W)$ or $(H, \frac{W}{l})$ where $l = 4$ is chosen for the S4 stage of the backbone. Reducing the receptive field to $\frac{1}{4}$th of the original but at the same time reducing the computational cost of the Attention mechanism, allowing for the use of self-attention in the S4 stage of the YOLO backbone.
    \item \textbf{Use of new ELAN block, A2C2f for accommodating Transformer modules:} YOLOv12 uses R-ELAN, a block very similar to VajraMerudandaBhag5 \cite{vajrav1_vayuai}. This is an ELAN block that does not split the feature map after the first 1x1 convolution and instead uses only 1 branch for the bottleneck blocks / inner blocks. The first R-ELAN like block was the VajraMerudandaBhag5 published in the VayuAI repository \cite{vajrav1_vayuai} in October 2024, and the R-ELAN \cite{tian2025yolov12attentioncentricrealtimeobject} blocks implemented in the neck of the YOLOv12's N, S and M models are identical to the VajraMerudandaBhag5. The novelty introduced in the YOLOv12 is the use of residual connections and layer scaling in the R-ELAN which is used in the S4 and S5 stages of the backbone when accommodating multiple Transformer modules for the purpose of stabilizing the training for L and X models. YOLOv12's L and X models use 8 Transformer modules in each A2C2f block both in the S4 and S5 stages of the backbone, resulting in a total of 16 Transformer modules. 
    \item \textbf{Use of Flash Attention for reduced latency:} YOLOv12 makes use of FlashAttention \cite{dao2022flashattention, dao2023flashattention2} which operates in half precision, uses memory optimizations and makes use of efficient GPU utilization to speed up the implementation of the Self-Attention mechanism. YOLOv12 relies on Flash Attention to maintain latency competitive with YOLOv11
\end{enumerate}

\section{Method}

The VajraV1 model architecture has the following characteristics:

\begin{enumerate}
    \item \textbf{Widened 3x3 Convolutions in the Primary Computational Block}
    \item \textbf{Use of a Parameter Efficient Computational Block}
    \item \textbf{Use of FLOP and Parameter Efficient Downsample Convolutions}
    \item \textbf{Use of Efficient Layer Aggregation Network for integrating multiple Transformer modules}
\end{enumerate}

\begin{figure}[H]
\centering
\includegraphics[width=0.9\textwidth]{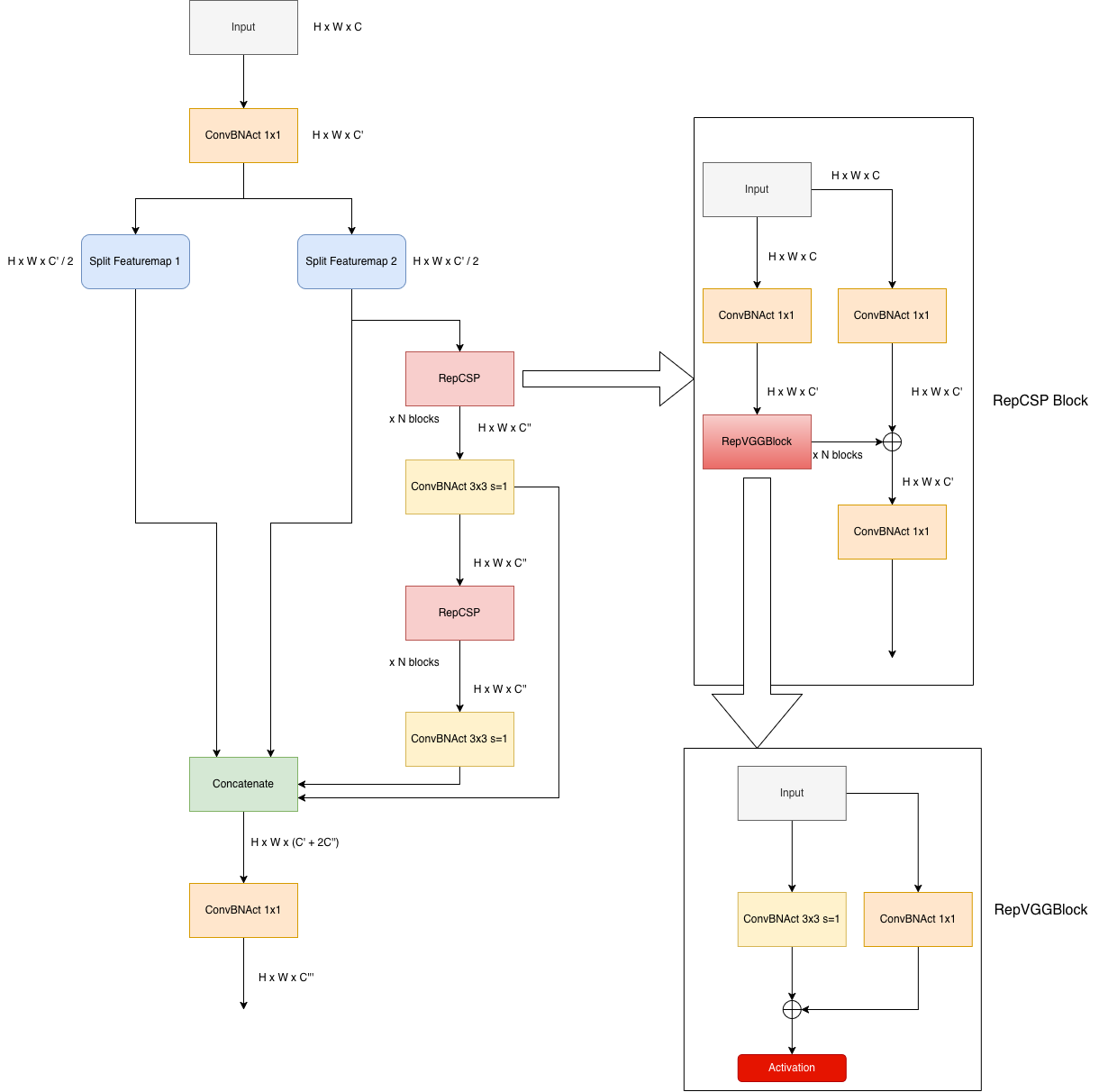}
\caption{VajraV1MerudandaX Block}
\label{fig:merudandaX}
\end{figure}

\begin{figure}[H]
\centering
\includegraphics[width=0.9\textwidth]{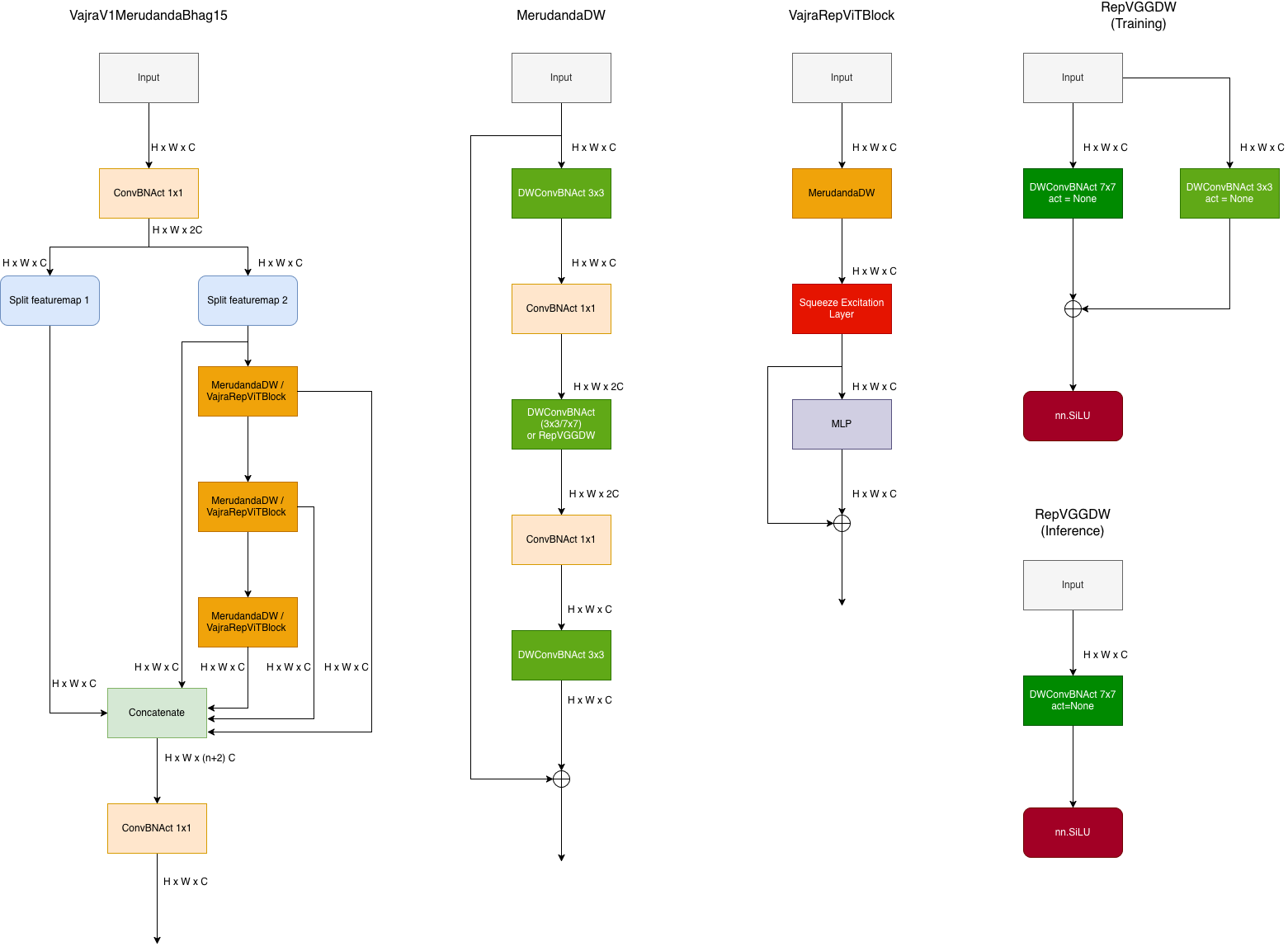}
\caption{VajraV1MerudandaBhag15 Block}
\label{fig:vajrav1_merudanda_bhag15}
\end{figure}

\begin{figure}[H]
\centering
\includegraphics[width=0.3\textwidth]{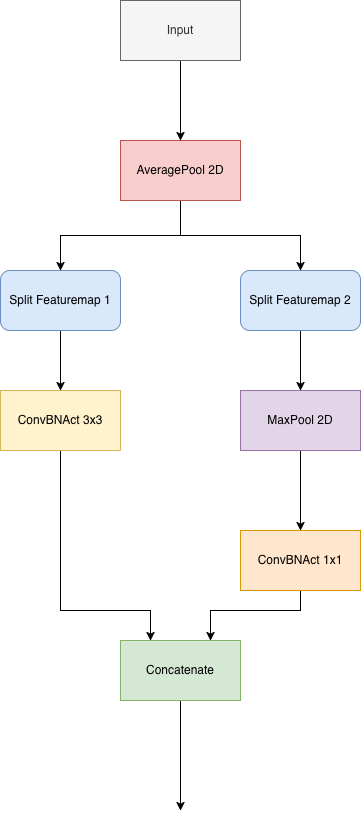}
\caption{ADown Module}
\label{fig:adown}
\end{figure}

A brief analysis of the architectural improvements of the VajraV1 compared to the recent YOLO models:

\begin{itemize}
    \item \textbf{VajraV1 vs YOLOv9:} With the help of a light classification head, FLOP efficient downsampling convolutions, parameter efficient computational blocks, VajraV1 is capable of integrating multiple Transformer modules (MHSA  + MLP modules) while still achieving lower latency than the YOLOv9 \cite{wang2024yolov9learningwantlearn} models. VajraV1 uses the VajraV1MerudandaX module as its primary computational block which is a modification of the RepNCSPELAN4 computational block of the YOLOv9 models modified by doubling the width of the RepVGGBlock used in the RepCSP module while halving the number of 3x3 convolutions compared to the YOLOv9. Additionally, the RepCSP module used in the VajraV1's VajraV1MerudandaX block is different from the RepNCSP module used in YOLOv9's RepNCSPELAN4 block as it uses a residual connection to join the two branches before the 1x1 projection convolution (as can be seen in Figure \ref{fig:merudandaX}) while YOLOv9's RepNCSP simply concatenates the two branches before the 1x1 projection convolution. It was observed that a combination of these methods allowed VajraV1 to outperform its YOLOv9 counterparts on the COCO dataset \cite{lin2015coco}.
    
    \item \textbf{VajraV1 vs YOLOv10:} VajraV1-Large and Xlarge integrate more Transformer blocks than their YOLOv10 \cite{wang2024yolov10realtimeendtoendobject} counterparts while managing lower latency in the case of VajraV1-Large. This increases the overall receptive field and enhances the global representation learning capability of VajraV1 compared to YOLOv10. VajraV1's parameter efficient VajraV1MerudandaBhag15 module (an ELAN that integrates MerudandaDW/VajraRepViTBlock modules) is adapted from YOLOv10's C2fCIB module (an ELAN that integrates the CIB block). The MerudandaDW block is identical to the CIB block used in YOLOv10 however, in the \textbf{S5 stage} of VajraV1's backbone VajraRepViTBlock module is chosen over the MerudandaDW module. VajraRepViTBlock is inspired by the RepViT \cite{wang2024repvitrevisitingmobilecnn} model architecture, it is a module that uses MerudandaDW as the "token mixer" and an MLP as the "channel mixer" (look at Figure \ref{fig:vajrav1_merudanda_bhag15}) similar to the strategy used in the RepViT. Integrating the VajraRepViTBlock in the VajraV1MerudandaBhag15 module in the \textbf{S5 stage} of the backbone and \textbf{P5 stage} of the neck of the VajraV1 yields better results than the MerudandaDW module alone, suggesting that \textbf{accommodating MLPs as channel mixers is beneficial to the model's overall representation learning}. Additionally, it was observed upon experimentation that the FLOP efficient downsample convolution, ADown used in the VajraV1 (given in Figure \ref{fig:adown}), though taken from the YOLOv9 \cite{wang2024yolov9learningwantlearn}, yields better results than the SCDown downsample convolution used in the YOLOv10 \cite{wang2024yolov10realtimeendtoendobject}.

    \item \textbf{VajraV1 vs YOLOv11:} The C3K2 block is the primary computational block used in the YOLOv11 \cite{yolo11_ultralytics} and is a GELAN similar to the VajraV1MerudandaX used in VajraV1 and the RepNCSPELAN4 used in YOLOv9. The VajraV1MerudandaX module has $\mathbf{2n + 2}$ 3x3 convolutions while the C3K2 has $\mathbf{4n}$ 3x3 convolutions where $\mathbf{n}$ is the number of blocks, for nano, small and medium models $\mathbf{n = 1}$, both the C3K2 and the VajraV1MerudandaX have the same number of 3x3 convolutions, while for the large and xlarge models $\mathbf{n = 2}$, the C3K2 has 2 more 3x3 convolutions. The 3x3 convolutions used in the VajraV1MerudandaX in the RepVGGBlock are given $\mathbf{2x}$ the width of each Bottleneck inside the C3K (inner block of the C3K2), resulting in an increase in the total FLOPs due to the width increase particularly in the P2 and P3 sections of the backbone. The increase in FLOPs has to be compensated by using FLOP efficient downsample convolutions like ADown and compact, parameter efficient computational blocks like VajraV1MerudandaBhag15 in certain sections of the backbone and the neck of the model. A combination of these techniques allows VajraV1 to achieve significantly better performance (mAP) on the COCO dataset \cite{lin2015coco} while achieving latency competitive with the YOLOv11 \cite{yolo11_ultralytics}.

    \item \textbf{VajraV1 vs YOLOv12:} 
    YOLOv12 \cite{tian2025yolov12attentioncentricrealtimeobject} uses a combination of C3K2 and A2C2f as its primary computational blocks. The A2C2f uses Area-Attention in the S4 stage of the backbone and Multi Head Self-Attention in the S5 stage of the backbone. The use of Area Attention allows the integration of Transformer Blocks in the backbone of the real-time object detector in stages of the backbone which receive higher resolution featuremaps (S4 stage receives featuremaps with total stride of 16 while S5 stage receives featuremaps with a total stride of 32). This is the main novelty introduced in the YOLOv12, before the introduction of Area-Attention, the use of Transformer blocks was limited to the S5 stage of the backbone.
    
    When the A2C2f is used in the neck of the model architecture, it is used without any Transformer modules and integrates the C3K (inner block of the C3K2) instead. The A2C2f uses the R-ELAN model architecture \textbf{which is a special case of the VajraMerudandaBhag5 (implemented in the VayuAI repo \cite{vajrav1_vayuai} in October 2024, months before the YOLOv12 paper was published)}. This can be observed in Figure \ref{fig:vajra_merudanda_bhag5}. The A2C2f module when used in the neck of the YOLOv12 \cite{tian2025yolov12attentioncentricrealtimeobject} model for the \textbf{N, S and M} models is used without any residual connection or layer scaling and is identical to the architecture of the VajraMerudandaBhag5. \\

    YOLOv12 \cite{tian2025yolov12attentioncentricrealtimeobject} uses the same width and depth scaling as the YOLOv11 \cite{yolo11_ultralytics} models for the bottleneck blocks and the inner blocks (C3K). When comparing the VajraV1MerudandaX module to A2C2f and C3K2 modules, it can be observed that each 3x3 convolution used in the VajraV1MerudandaX module is 2x as wide. For the Nano, Small, Medium models VajraV1MerudandaX block uses the same number of 3x3 convolutions as A2C2f and C3K2. But for Large and Xlarge models VajraV1MerudandaX is two 3x3 convolutions short when compared to the A2C2f and C3K2 blocks.\\

    It was observed that increasing the width of each of the bottleneck blocks with no compromise in the number of 3x3 convolutions per block gives VajraV1's \textbf{Nano, Small and Medium} models an upper hand over their YOLOv12 counterparts. However, the VajraV1-Large performs at-par or worse than the YOLOv12-L. The \textbf{VajraV1-Xlarge} on the other hand has a significantly larger receptive field than its YOLOv12 \cite{tian2025yolov12attentioncentricrealtimeobject} and YOLOv11 \cite{yolo11_ultralytics} counterparts and achieves better performance on the COCO \cite{lin2015coco} benchmarks.\\

    \begin{figure}[H]
    \centering
    \includegraphics[width=0.9\textwidth]{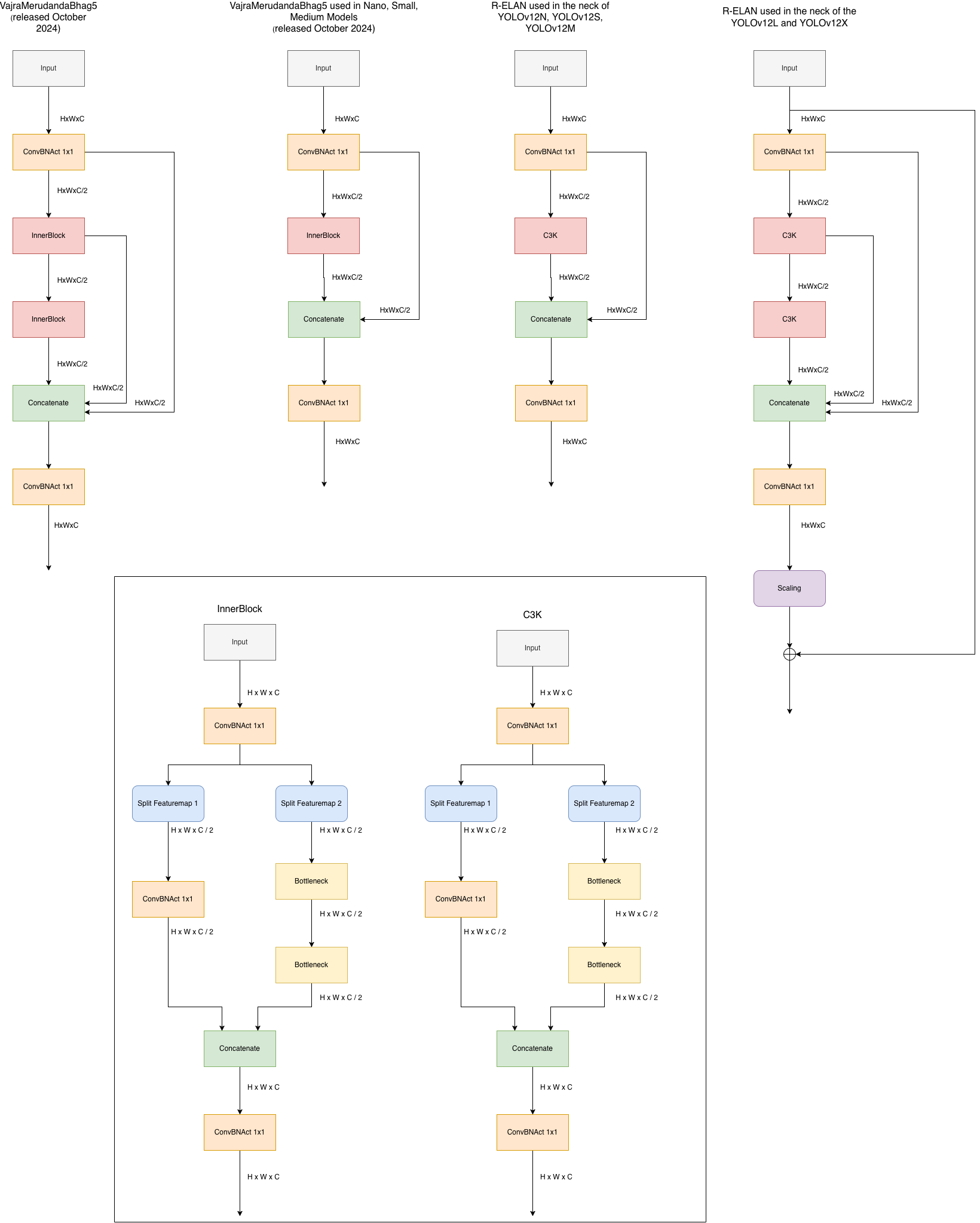}
    \caption{VajraMerudandaBhag5 vs YOLOv12's R-ELAN. It can be observed that the R-ELAN (A2C2f Block) \cite{tian2025yolov12attentioncentricrealtimeobject} used in the neck of the YOLOv12N, YOLOv12S and YOLOv12M is identical to the VajraMerudandaBhag5 Block introduced in the VayuAI repository \cite{vajrav1_vayuai} in October 2024, months before the YOLOv12 paper \cite{tian2025yolov12attentioncentricrealtimeobject} was published.}
    \label{fig:vajra_merudanda_bhag5}
    \end{figure}
    
\end{itemize}

\subsection{Primary Computational Block - VajraV1MerudandaX}

VajraMerudandaX is the primary computational block of the VajraV1 model and is used throughout the backbone and the neck of the model architecture. The VajraMerudandaX block is inspired by the GELAN architecture utilized in the YOLOv9 \cite{wang2024yolov9learningwantlearn} models, specifically the RepNCSPELAN4 module. However, compared to YOLOv9, VajraV1 uses a different architecture for the RepCSP module that is used inside the VajraV1MerudandaX block compared to the RepNCSP module used inside YOLOv9's RepNCSPELAN4 block. VajraV1MerudandaX is inspired by and is identical to the modified RepNCSPELAN4 module used in the neck of the D-FINE DETR \cite{peng2024dfineredefineregressiontask} and DEIM DETR \cite{huang2025deimdetrimprovedmatching}.

\subsubsection{RepCSP Block}
The RepCSP module integrates multiple RepVGGBlock \cite{ding2021repvggmakingvggstyleconvnets} modules inside it. It consists of two branches, the first branch consisting of a 1x1 convolution followed by a series of RepVGGBlock modules, the second branch consists of a 1x1 convolution which is joined with the first branch through a residual connection before the final 1x1 convolution. The main branch which houses the RepVGGBlock modules generates more semantic information with a large receptive field while the second branch preserves more fine-grained spatial information with a small receptive field.

\subsubsection{RepVGGBlock}
The RepVGGBlock \cite{ding2021repvggmakingvggstyleconvnets} module consists of two branches, the first being a 3x3 convolution and the second being a 1x1 convolution, both implemented with BatchNorm. Both branches are joined together with the help of a residual connection. During inference however, the RepVGGBlock is re-parameterized and the 1x1 convolution's weights and biases are fused with those of the 3x3 convolution, resulting in zero latency overhead for the module when compared to a regular 3x3 convolution. While enjoying the advantages of a 1x1 convolution identity branch during training such as providing linear channel mixing, providing the model with linear shortcuts and providing low receptive field information.

\subsection{Parameter Efficient Computational Block - VajraV1MerudandaBhag15}

The VajraV1MerudandaBhag15 utilizes the MerudandaDW block (which is identical to the CIB block used in the YOLOv10 \cite{wang2024yolov10realtimeendtoendobject}) and the VajraRepViTBlock which is a novel computational block inspired by the RepViT \cite{wang2024repvitrevisitingmobilecnn} model architecture. The MerudandaDW and the Squeeze Excitation Layer \cite{hu2019squeezeandexcitationnetworks} are used as the \textbf{token mixer} while the MLP \cite{tolstikhin2021mlpmixer} acts as the \textbf{channel mixer}. The VajraV1MerudandaBhag15 block is used identically to the C2fCIB block of the YOLOv10 \cite{wang2024yolov10realtimeendtoendobject} model architecture following the rank-guided block design philosophy. \\

\textbf{Large Kernel Convolution} - A depthwise 7x7 convolution is utilized in the MerudandaDW module for the nano and small models of VajraV1. This too is inspired by the YOLOv10, the accuracy improves when 3x3 convolutions are replaced by 7x7 convolutions in the MerudandaDW in the final stage of the neck (P5) in the \textbf{Nano} model and the final stage of both the backbone (S5) and the neck (P5) in the \textbf{Small} model. However, unlike YOLOv10 when the VajraV1MerudandaBhag15 module is integrated in the S5 stage of the VajraV1 backbone, the VajraRepViTBlock is used instead of MerudandaDW / CIB. 

\subsection{Attention Block - VajraV1AttentionBhag6}

\begin{figure}[H]
\centering
\includegraphics[width=0.9\textwidth]{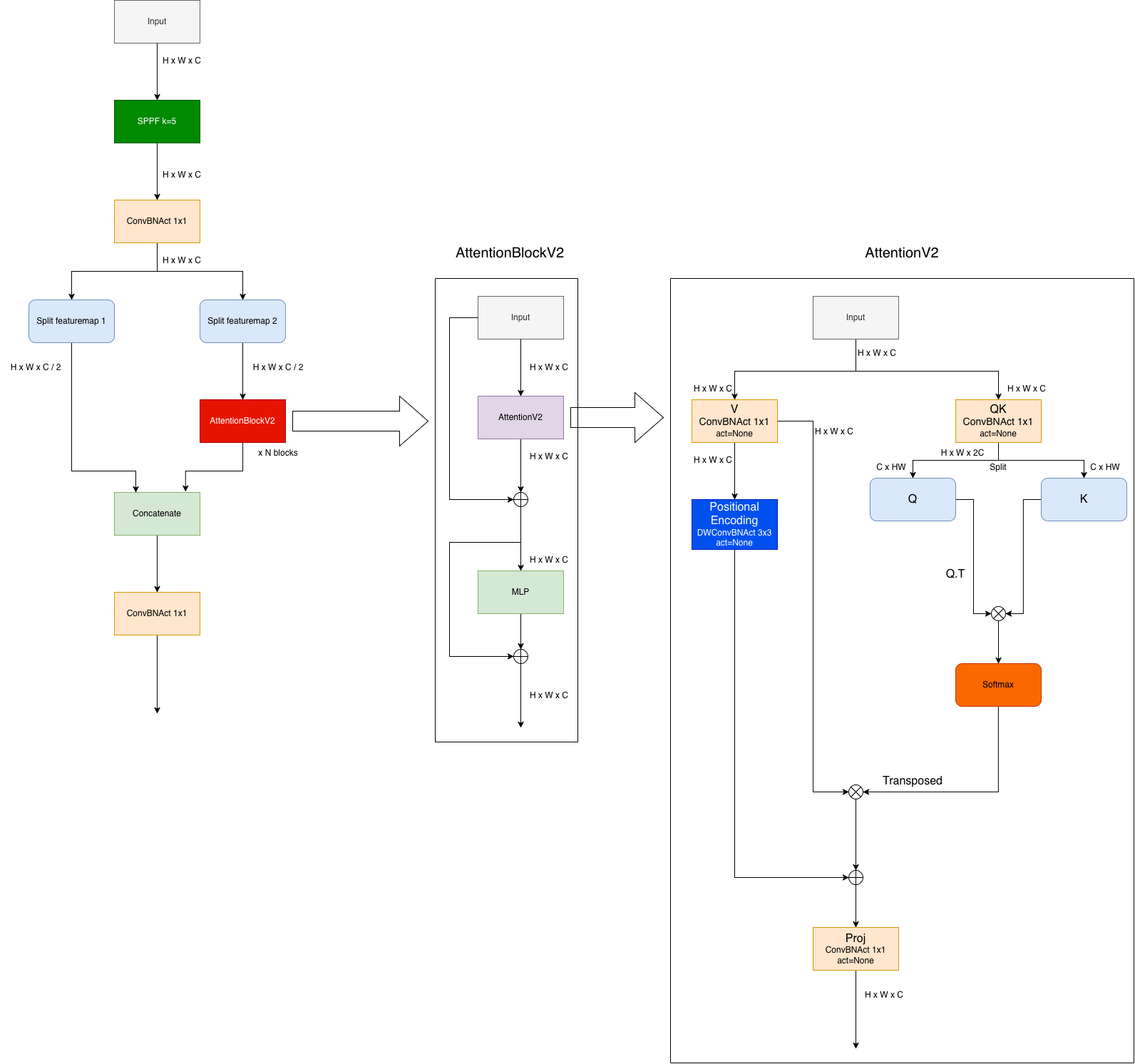}
\caption{VajraV1AttentionBhag6 Block}
\label{fig:vajrav1_attention_bhag6}
\end{figure}

The VajraV1AttentionBhag6 is similar to the C2PSA block used in YOLOv11 \cite{yolo11_ultralytics}. This module implements a convolutional block with attention mechanisms to enhance feature extraction and processing capabilities. It accommodates multiple AttentionBlockV2 blocks which is a Transformer block, each of which consists of Multi-Head Self-Attention (MHSA) modules and MLPs. Unlike the Positional Self-Attention module used in the YOLOv10 the VajraV1AttentionBhag6 block integrates two Transformer modules for the \textbf{large and xlarge} compared to the singular transformer module used in the YOLOv10's Positional Self-Attention. The VajraV1AttentionBhag6 block is used in the last stage of the backbone of the VajraV1 model (S5) and it is therefore applied only on the lowest resolution featuremap. Given an input featuremap $x \in \mathbb{R}^{B \times C \times H \times W}$ The computation process of the VajraV1AttentionBhag6 block is as follows:

\[x' = SPPF(x)\]
\[x'' = ConvBNAct^{1\times1}(x')\]
\[x_1'', x_2'' = Split(x'')\]
\[x_2''' = AttentionBlockV2_{\times N}(x_2'')\]
\[x'''' = Concat(x_1'', x_2''')\]
\[Output = ConvBNAct^{1\times1}(x'''')\]

\subsubsection{AttentionV2}

AttentionV2 is an efficient multi-head self-attention \cite{vaswani2017attention} mechanism designed for high-resolution feature maps in convolutional neural networks. It replaces standard dot-product attention with a lightweight, convolution-based formulation that maintains performance while reducing computational overhead.\\

The module takes an input featuremap $x \in \mathbb{R}^{B \times C \times H \times W}$ and computes multi-head self-attention as follows:

\begin{enumerate}
    \item Three parallel $1\!\times\!1$ convolutions are applied:
    \begin{itemize}
        \item $\texttt{qk}(\cdot)$: maps $C \to 2C$ channels to jointly produce query and key features,
        \item $\texttt{v}(\cdot)$: maps $C \to C$ to produce value features,
        \item $\texttt{proj}(\cdot)$: final $1\!\times\!1$ projection back to $C$ channels.
    \end{itemize}
    \item An additional depthwise $3\!\times\!3$ convolution with group cardinality $C$ generates a spatially-aware \textbf{positional encoding} (used in \cite{liu2021swin, wang2024yolov10realtimeendtoendobject, yolo11_ultralytics, tian2025yolov12attentioncentricrealtimeobject}) term from the value features.
\end{enumerate}

The output of $\texttt{qk}(x)$ is reshaped to $(B, \text{num\_heads}, 2 \cdot d_\text{head}, H\!\times\!W)$ and split equally along the channel dimension into queries $Q$ and keys $K$, each of size $d_\text{head}$. The value branch is reshaped to $(B, \text{num\_heads}, d_\text{head}, H\!\times\!W)$.\\

Attention scores are computed using standard scaled dot-product attention:
\[
\text{Attention}(Q, K, V) = \text{softmax}\left( \frac{Q K^T}{\sqrt{d_{\text{head}}}} \right) V
\]
where $d_{\text{head}} = C / \text{num\_heads}$ is the dimension per head.\\

For maximum efficiency on modern GPUs, the implementation can also utilize \textbf{FlashAttention-2} \cite{dao2022flashattention, dao2023flashattention2} (if available and input is on CUDA), operating in half-precision and using the optimal memory layout $(B, \text{heads}, N, d_{\text{head}})$. When FlashAttention is unavailable, a manually optimized PyTorch kernel is used. \textbf{Note: VajraV1 models have not yet been trained using FlashAttention}.\\

A key design choice is the injection of \textbf{relative positional bias} via a learned depthwise $3\!\times\!3$ convolution applied to the value features. This positional encoding is added (residual-style) to the attention output \textbf{before} the final projection:
\[
x_{\text{out}} = \texttt{proj}\left( \text{Attention}(Q, K, V) + \texttt{positional\_encoding}(V) \right)
\]

Unlike standard Transformer-based attention \cite{vaswani2017attention}, \textbf{all} LayerNorm \cite{ba2016layernormalization} layers are deliberately replaced with BatchNorm \cite{ioffe2015batchnormalizationacceleratingdeep} to reduce latency in convolutional backbones, particularly when processing large feature maps. This substitution has been shown empirically to maintain accuracy while offering speedup in wall-clock time on typical vision backbones.\\

This combination of shared QK convolution, depthwise positional encoding, FlashAttention support, and BatchNorm makes AttentionV2 one of the fastest drop-in attention mechanisms for high-resolution CNNs and hybrid CNN-Transformer architectures.

\subsection{FLOP efficient downsample convolution - ADown}

The ADown module introduced in the YOLOv9 \cite{wang2024yolov9learningwantlearn} integrates average pooling and max pooling with a 3x3 downsample convolution to capture diverse feature representations. This is done in a manner that is FLOP and parameter efficient by halving the width of the 3x3 downsample convolution. Given an input featuremap $x \in \mathbb{R}^{B \times C \times H \times W}$, the computation process is as follows:

\[x' = AvgPool(x),\]
\[x_1', x_2' = Split(x'),\]
\[x_1'' = ConvBNAct^{3\times3} (x_1'),\]
\[x_2'' = ConvBNAct^{1\times1} (MaxPool(x_2')),\]
\[Output = Concat(x_1'', x_2'').\]

\paragraph{Computational Cost:}
Let the input feature map be $x \in \mathbb{R}^{B \times C \times H \times W}$ and the output have $C_{\text{out}}$ channels. A standard $3{\times}3$ stride-2 convolution has a computational cost of
\[
\text{MAC}_{3\times3\text{-std}} 
= B H W \cdot \frac{9}{4} \, C \, C_{\text{out}}.
\]

In contrast, the ADown module first applies $2{\times}2$ average pooling with stride~1, then splits the channels into two equal parts. The first branch applies a $3{\times}3$ stride-2 convolution mapping $(C/2) \to (C_{\text{out}}/2)$ channels, resulting in
\[
\text{MAC}_{\text{branch-1}}
= B H W \cdot \frac{9}{16} \, C \, C_{\text{out}}.
\]
The second branch performs $3{\times}3$ max pooling (cost negligible) followed by a $1{\times}1$ convolution mapping $(C/2) \to (C_{\text{out}}/2)$ channels:
\[
\text{MAC}_{\text{branch-2}}
= B H W \cdot \frac{1}{16} \, C \, C_{\text{out}}.
\]
Thus, the total cost of ADown is
\[
\text{MAC}_{\text{ADown}}
= B H W \cdot \frac{5}{8} \, C \, C_{\text{out}}.
\]
Compared to the standard downsampling convolution, the relative cost is
\[
\frac{\text{MAC}_{\text{ADown}}}{\text{MAC}_{3\times3\text{-std}}}
= \frac{5/8}{9/4}
= \frac{5}{18}
\approx 0.28.
\]

\noindent
\textbf{Therefore, ADown uses only 27.7\% of the FLOPs and parameters of a standard $3{\times}3$ stride-2 convolution}, making it approximately \textbf{3.6$\times$ more efficient} while preserving representational diversity through mixed pooling and dual-branch processing.

\section{Experiments}

VajraV1 models were trained on the MS COCO 2017 Dataset \cite{lin2015coco} for \textbf{object detection, instance segmentation, and pose estimation (keypoint detection)} tasks. The Box mAP, Mask mAP, Pose mAP, and TensorRT10 FP16 latency on NVIDIA RTX-4090 can be visualized in the following tables: Table \ref{tab:results_detection}, Table \ref{tab:results_seg}, Table \ref{tab:results_pose}.

\subsection{Experimental Setup}
\begin{itemize}
    \item \textbf{GPU}: 8x NVIDIA RTX-4090
    \item \textbf{Batch Size}: 128
    \item \textbf{Epochs}: 600 for the MS COCO \cite{lin2015coco} Detection and Segmentation Datasets, 1000 for the MS COCO \cite{lin2015coco} Pose Estimation Dataset.
    \item \textbf{Optimizer}: SGD with Nesterov Momentum
    \item \textbf{Initial LR}: $1 \times 10^{-2}$
    \item \textbf{Final LR}: $1 \times 10^{-4}$
    \item \textbf{Momentum}: 0.937
    \item \textbf{Weight Decay}: $5 \times 10^{-4}$
    \item \textbf{Warmup Epochs}: 3
    \item \textbf{Warmup Momentum}: 0.8
\end{itemize}

\subsection{Evaluation Protocol:}
All VajraV1 variants are evaluated on the COCO 2017 \cite{lin2015coco} validation set using the official \textbf{pycocotools API}, the standard evaluation toolkit widely adopted in the object detection community. To ensure reproducibility and fair comparison, core components such as data augmentation (Mosaic \cite{bochkovskiy2020yolov4optimalspeedaccuracy, wang2022yolov7trainablebagoffreebiessets}, Mixup \cite{zhang2018mixupempiricalriskminimization}, Copy Paste \cite{ghiasi2021simplecopypastestrongdata}), non-maximum suppression, and training pipelines are adapted from the \textbf{Ultralytics YOLO repository} \cite{yolo11_ultralytics}, which has been rigorously validated by the community. This methodology guarantees that the reported mAP values are directly comparable to previously published YOLO models (YOLOv11 \cite{yolo11_ultralytics}, YOLOv12 \cite{tian2025yolov12attentioncentricrealtimeobject}), eliminating potential inconsistencies arising from differing evaluation setups.\\

\begin{table}[H]
  \centering
  \begin{tabular}{lccccc}
    \toprule
    Model & Params (M) & FLOPs (B) & Box $AP^{val}_{50-95}$ (\%) & Latency (ms/img) \\
    \midrule
    YOLOv10-N \cite{wang2024yolov10realtimeendtoendobject} & 2.3 & 6.7 & 39.5 & 1.1 \\
    YOLOv11-N \cite{yolo11_ultralytics} & 2.6 & 6.5 & 39.5 & 0.9 \\
    YOLOv12-N \cite{tian2025yolov12attentioncentricrealtimeobject} & 2.5 & 6.0 & 40.4 & 0.9 \\
    YOLOv13-N \cite{lei2025yolov13realtimeobjectdetection} & 2.5 & 6.4 & 41.6 & 1.25 \\
    \textbf{VajraV1-Nano} & 3.78 & 13.7 & \textbf{44.3} & 1.1 \\
    \midrule
    YOLOv10-S \cite{wang2024yolov10realtimeendtoendobject} & 7.2 & 21.6 & 46.8 & 1.1 \\
    YOLOv11-S \cite{yolo11_ultralytics} & 9.4 & 21.5 & 47 & 1.1 \\
    YOLOv12-S \cite{tian2025yolov12attentioncentricrealtimeobject} & 9.1 & 19.4 & 47.6 & 1.0 \\
    YOLOv13-S \cite{lei2025yolov13realtimeobjectdetection} & 9.0 & 20.8 & 48.0 & 1.27\\
    \textbf{VajraV1-Small} & 11.58 & 47.9 & \textbf{50.4} & 1.1 \\
    \midrule
    YOLOv10-B \cite{wang2024yolov10realtimeendtoendobject} & 19.1 & 92.0 & 52.7 & 1.7\\
    YOLOv11-M \cite{yolo11_ultralytics} & 20.1 & 68.0 & 51.5 & 1.4 \\
    YOLOv12-M \cite{tian2025yolov12attentioncentricrealtimeobject} & 19.6 & 59.8 & 52.5 & 1.3 \\
    \textbf{VajraV1-Medium} & 20.29 & 94.5 & \textbf{52.7} & 1.5\\
    \midrule
    YOLOv10-L \cite{wang2024yolov10realtimeendtoendobject} & 24.4 & 120.3 & 53.4 & 1.9\\
    YOLOv11-L \cite{yolo11_ultralytics} & 25.3 & 86.9 & 53.4 & 1.7\\
    YOLOv12-L \cite{tian2025yolov12attentioncentricrealtimeobject} & 26.5 & 82.4 & \textbf{53.8} & 2.0\\
    YOLOv13-L \cite{lei2025yolov13realtimeobjectdetection} & 27.6 & 88.4 & 53.4 & 2.36\\
    \textbf{VajraV1-Large} & 24.63 & 115.2 & 53.7 & 1.8\\
    \midrule
    YOLOv10-X \cite{wang2024yolov10realtimeendtoendobject} & 29.5 & 160.4 & 54.4 & 2.3\\
    YOLOv11-X \cite{yolo11_ultralytics} & 56.9 & 194.9 & 54.7 & 2.5\\
    YOLOv12-X \cite{tian2025yolov12attentioncentricrealtimeobject} & 59.3 & 184.6 & 55.4 & 2.4\\
    YOLOv13-X \cite{lei2025yolov13realtimeobjectdetection} & 64.0 & 199.2  & 54.8 & 3.1\\
    \textbf{VajraV1-Xlarge} & 72.7 & 208.3 & \textbf{56.2} & 3.2\\
    \bottomrule
  \end{tabular}
  \caption{Comparison on COCO Validation Set for Object Detection. Note that the Latency is measured using TensorRT10 with FP16 inference on an NVIDIA RTX-4090 GPU. \textbf{For YOLOv13 models, latency values are taken directly from the original paper and were not independently re-evaluated under the same TensorRT setup.}}
  \label{tab:results_detection}
\end{table}

\begin{table}[H]
  \centering
  \begin{tabular}{lccccc}
    \toprule
    Model & Params (M) & FLOPs (B) & Mask $AP^{val}_{50-95}$ (\%) & Latency (ms/img) \\
    \midrule
    YOLOv11-N \cite{yolo11_ultralytics} & 2.9 & 9.7 & 32.0 & 1.1 \\
    YOLOv12-N \cite{tian2025yolov12attentioncentricrealtimeobject} & 2.8 & 9.9 & 32.8 & 1.1 \\
    \textbf{VajraV1-Nano} & 4.03 & 17.6 & \textbf{35.8} & 1.2 \\
    \midrule
    YOLOv11-S \cite{yolo11_ultralytics} & 10.1 & 33.0 & 37.8 & 1.1\\
    YOLOv12-S \cite{tian2025yolov12attentioncentricrealtimeobject} & 9.8 & 33.4 & 38.6 & 1.1\\
    \textbf{VajraV1-Small} & 12.23 & 61.9 & \textbf{40.5} & 1.2\\
    \midrule
    YOLOv11-M \cite{yolo11_ultralytics} & 22.4 & 113.2 & 41.5 & 1.5\\
    YOLOv12-M \cite{tian2025yolov12attentioncentricrealtimeobject} & 21.9 & 115.1 & 42.3 & 1.6\\
    \textbf{VajraV1-Medium} & 22.6 & 149.9 & \textbf{42.3} & 1.7\\
    \midrule
    YOLOv11-L \cite{yolo11_ultralytics} & 27.6 & 132.2 & 42.9 & 2.0\\
    YOLOv12-L \cite{tian2025yolov12attentioncentricrealtimeobject} & 28.8 & 137.7 & \textbf{43.2} & 2.0\\
    \textbf{VajraV1-Large} & 26.93 & 170.6 & 43.1 & 2.0\\
    \midrule
    YOLOv11-X \cite{yolo11_ultralytics} & 62.1 & 296.4 & 43.8 & 2.9\\
    YOLOv12-X \cite{lei2025yolov13realtimeobjectdetection} & 64.5 & 308.7 & 44.2 & 2.9\\
    \textbf{VajraV1-Xlarge} & 75 & 278.1 & \textbf{44.5} & 3.4\\
    \bottomrule
  \end{tabular}
  \caption{Comparison on COCO Validation Set for Instance Segmentation. Note that the Latency is measured using TensorRT10 with FP16 inference on an NVIDIA RTX-4090 GPU}
  \label{tab:results_seg}
\end{table}

\begin{table}[H]
  \centering
  \begin{tabular}{lccccc}
    \toprule
    Model & Params (M) & FLOPs (B) & Pose $AP^{val}_{50-95}$ (\%) & Latency (ms/img) \\
    \midrule
    YOLOv11-N \cite{yolo11_ultralytics} & 2.9 & 7.4 & 50.0 & 1.0 \\
    \textbf{VajraV1-Nano} & 4.07 & 14.8 & \textbf{56.4} & 1.1 \\
    \midrule
    YOLOv11-S \cite{yolo11_ultralytics} & 9.9 & 23.1 & 58.5 & 1.1 \\
    \textbf{VajraV1-Small} & 12.07 & 49.6 & \textbf{65.0} & 1.4\\
    \midrule
    YOLOv11-M \cite{yolo11_ultralytics} & 20.9 & 71.4 & 64.9 & 1.5 \\
    \textbf{VajraV1-Medium} & 21.15 & 98.4 & \textbf{68.5} & 1.8 \\
    \midrule
    YOLOv11-L \cite{yolo11_ultralytics} & 26.1 & 90.3 & 66.1 & 2.0 \\
    \textbf{VajraV1-Large} & 25.49 & 118.9 & \textbf{69.5} & 2.1 \\
    \midrule
    YOLOv11-X \cite{yolo11_ultralytics} & 58.8 & 202.8 & 69.5 & 2.8 \\
    \textbf{VajraV1-Xlarge} & 73.56 & 226.5 & \textbf{71.5} & 3.7\\
    \bottomrule
  \end{tabular}
  \caption{Comparison on COCO Validation Set for Pose Estimation. Note that the Latency is measured using TensorRT10 with FP16 inference on an NVIDIA RTX-4090 GPU}
  \label{tab:results_pose}
\end{table}

\subsection{Comparison with State-of-the-arts}

\subsubsection{COCO Detection Benchmark}

\begin{itemize}
    \item \textbf{N-scale models:} VajraV1-Nano outperforms YOLOv13-N \cite{lei2025yolov13realtimeobjectdetection}, YOLOv12-N \cite{tian2025yolov12attentioncentricrealtimeobject}, YOLOv11-N \cite{yolo11_ultralytics}, YOLOv10-N \cite{wang2024yolov10realtimeendtoendobject} by \textbf{2.7\%, 3.7\%, 4.8\%, and 4.8\%} respectively.
    \item \textbf{S-scale models:} VajraV1-Small outperforms YOLOv13-S \cite{lei2025yolov13realtimeobjectdetection}, YOLOv12-S \cite{tian2025yolov12attentioncentricrealtimeobject}, YOLOv11-S \cite{yolo11_ultralytics}, YOLOv10-S \cite{wang2024yolov10realtimeendtoendobject}, YOLOv9-S \cite{wang2024yolov9learningwantlearn} by \textbf{2.4\%, 2.8\%, 3.4\%, 3.6\%, and 3.6\%} respectively. 
    \item \textbf{M-scale models:} VajraV1-Medium outperforms YOLOv12-M \cite{tian2025yolov12attentioncentricrealtimeobject}, YOLOv11-M \cite{yolo11_ultralytics}, YOLOv10-M \cite{wang2024yolov10realtimeendtoendobject} and YOLOv9-M \cite{wang2024yolov9learningwantlearn} by \textbf{0.2\%, 1.2\%, 1.4\%, 1.3\%} respectively. \textbf{Note that the VajraV1-Medium achieves the same $AP_{50-95}^{val}$ as the YOLOv10-B \cite{wang2024yolov10realtimeendtoendobject}} which is closer to the VajraV1-Medium in terms of FLOPs than the YOLOv10-M.
    \item \textbf{L-scale models:} VajraV1-Large performs slightly worse on the COCO detection benchmark than the YOLOv12-L \cite{tian2025yolov12attentioncentricrealtimeobject} (\textbf{53.7\% vs 53.8\%}). However, VajraV1-Large outperforms the YOLOv13-L \cite{lei2025yolov13realtimeobjectdetection}, YOLOv11-L \cite{yolo11_ultralytics} and YOLOv10-L \cite{wang2024yolov10realtimeendtoendobject} by \textbf{0.3\%} and the YOLOv9-C \cite{wang2024yolov9learningwantlearn} by \textbf{0.7\%}.
    \item \textbf{X-scale models:} VajraV1-Xlarge outperforms YOLOv13-X \cite{lei2025yolov13realtimeobjectdetection}, YOLOv12-X \cite{tian2025yolov12attentioncentricrealtimeobject}, YOLOv11-X \cite{yolo11_ultralytics}, YOLOv10-X \cite{wang2024yolov10realtimeendtoendobject}, YOLOv9-E \cite{wang2024yolov9learningwantlearn} by \textbf{1.4\%, 0.8\%, 1.5\%, 1.8\%, and 0.6\%} respectively, outperforming all models in the YOLO family on the COCO detection benchmark.
\end{itemize}

\subsubsection{COCO Segmentation Benchmark}
For the COCO Segmentation Benchmark, the Mask mAP metric is utilized. VajraV1's performance is compared to the YOLOv12 and YOLOv11 as these are the only other YOLO models trained on the COCO Segmentation Benchmark.
\begin{itemize}
    \item \textbf{N-scale models:} VajraV1-Nano achieves a \textbf{Mask mAP of 35.8\%}, outperforming the YOLOv12-N \cite{tian2025yolov12attentioncentricrealtimeobject}, YOLOv11-N \cite{yolo11_ultralytics} by \textbf{3\%, and 3.8\%} respectively.
    \item \textbf{S-scale models:} VajraV1-Small achieves a \textbf{Mask mAP of 40.5} outperforms YOLOv12-S \cite{tian2025yolov12attentioncentricrealtimeobject}, YOLOv11-S \cite{yolo11_ultralytics} by \textbf{1.9\% and 2.7\%} respectively. 
    \item \textbf{M-scale models:} VajraV1-Medium achieves a \textbf{Mask mAP of 42.3} performing at-par with YOLOv12-M \cite{tian2025yolov12attentioncentricrealtimeobject} and outperforming YOLOv11-M \cite{yolo11_ultralytics} by \textbf{0.8\%}.
    \item \textbf{L-scale models:} VajraV1-Large achives a \textbf{Mask mAP of 43.1} outperforming the YOLOv11-L \cite{yolo11_ultralytics} by \textbf{0.2\%} while just falling short of the YOLOv12-L's \cite{tian2025yolov12attentioncentricrealtimeobject} \textbf{Mask mAP of 43.2} by \textbf{0.1\%}.
    \item \textbf{X-scale models:} VajraV1-Xlarge outperforms its YOLOv12 \cite{tian2025yolov12attentioncentricrealtimeobject} and YOLOv11 \cite{yolo11_ultralytics} counterparts achieving \textbf{Mask mAP of 44.5\%} outperforming the YOLOv12-X \cite{tian2025yolov12attentioncentricrealtimeobject} and YOLOv11-X \cite{yolo11_ultralytics} by \textbf{0.3\% and 1.7\%} respectively
\end{itemize}

\subsubsection{COCO Pose Estimation Benchmark}
For the COCO Pose Estimation Benchmark, the Pose mAP metric is utilized. VajraV1's performance is compared to the YOLOv11. VajraV1 and YOLOv11 are the only two YOLO models trained on the COCO Pose Estimation Benchmark so far.
\begin{itemize}
    \item \textbf{N-scale models:} VajraV1-Nano outperforms YOLOv11-N \cite{yolo11_ultralytics} by \textbf{6.4\%} while achieving comparable latency of 1.1 ms/img using TensorRT10 with FP16 inference on an NVIDIA RTX-4090 GPU.
    \item \textbf{S-scale models:} VajraV1-Small outperforms YOLOv11-S \cite{yolo11_ultralytics} by \textbf{6.5\%} while getting slightly worse latency of 1.4 ms/img compared to 1.1 ms/img of the YOLOv11-S.
    \item \textbf{M-scale models:} VajraV1-Medium outperforms YOLOv11-M \cite{yolo11_ultralytics} by \textbf{3.6\%} while being slower by 0.3 ms/img compared to the YOLO11-M.
    \item \textbf{L-scale models:} VajraV1-Large outperfroms YOLOv11-L \cite{yolo11_ultralytics} by \textbf{3.4\%} and \textbf{achieves the same Pose mAP as the YOLOv11-X of 69.5\%} while achieving latency comparable to the YOLOv11-L of \textbf{2.1 ms/img compared to YOLOv11-L's 2.0 ms/img}. \\
    
    \textbf{VajraV1-Large} on the Pose Estimation benchmark achieves \textbf{Pose mAP equal to the YOLOv11-X \cite{yolo11_ultralytics}} while achieving \textbf{latency comparable to the YOLOv11-L \cite{yolo11_ultralytics}}. This is truly a testament to how capable the VajraV1 models are.
    \item \textbf{X-scale models:} VajraV1-Xlarge achieves a Pose $AP^{val}_{50-95}\%$ of \textbf{71.5\%} compared to YOLOv11-X's \cite{yolo11_ultralytics} \textbf{69.5\%}, outperforming it by \textbf{2\%} while being 0.9 ms/img slower than it.
\end{itemize}

\subsection{Heatmap Visualization}

\begin{figure}[H]
\centering
\includegraphics[width=\textwidth]{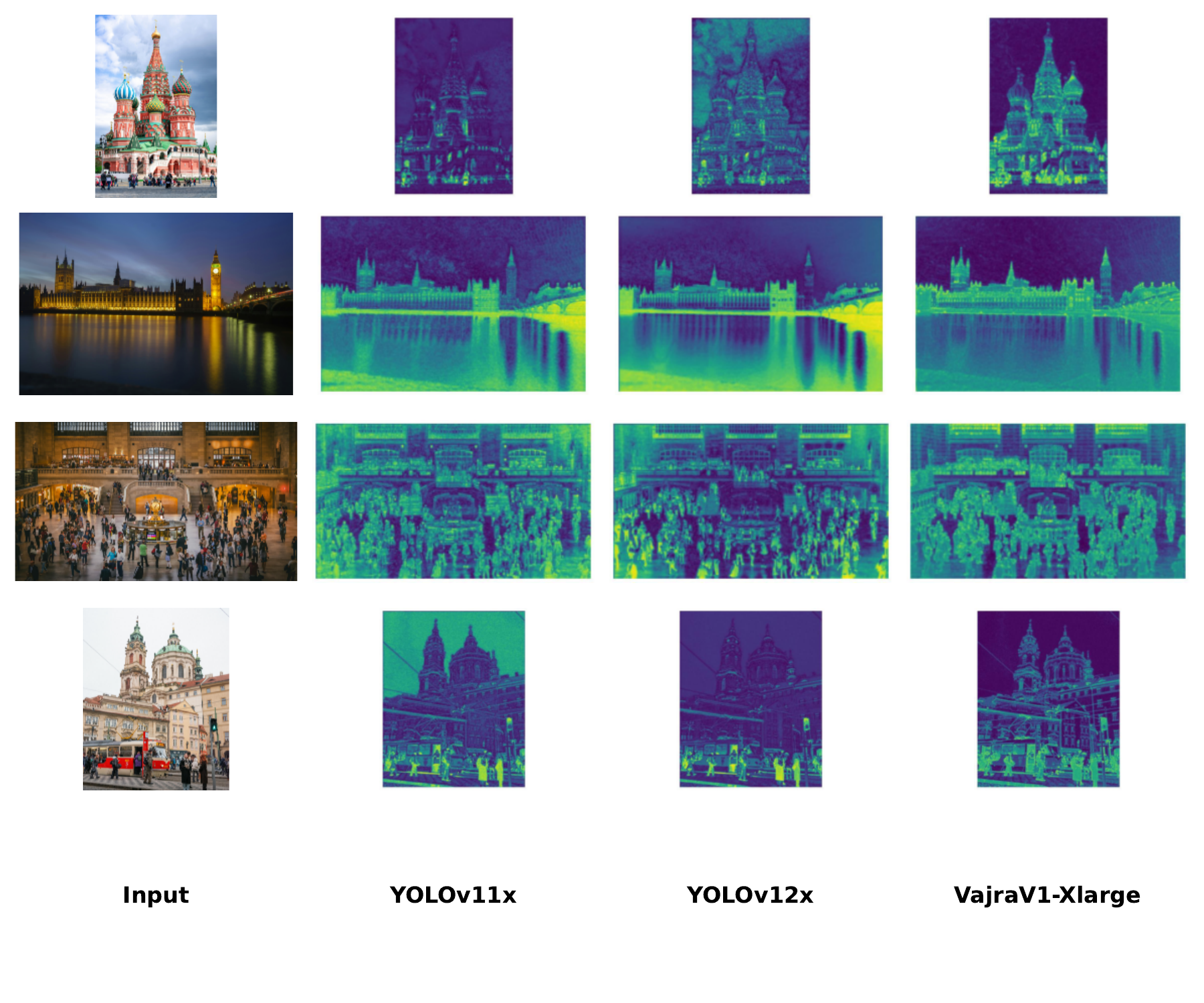}
\caption{Comparison of heat maps between YOLOv11X \cite{yolo11_ultralytics}, YOLOv12X \cite{tian2025yolov12attentioncentricrealtimeobject} and the new VajraV1-Xlarge. It can be observed that the heat maps of the VajraV1Xlarge have better foreground-background separation and are less noisy when compared to the heatmaps of its YOLOv11 and YOLOv12 counterparts.}
\label{fig:heatmaps}
\end{figure}

Figure \ref{fig:heatmaps} compares the heat maps of the VajraV1-Xlarge with those of the YOLOv12X \cite{tian2025yolov12attentioncentricrealtimeobject} and YOLOv11X \cite{yolo11_ultralytics}. These heatmaps are extracted from the S4 stage of the backbone (stride of 16) of each model. It can be observed that VajraV1-Xlarge produces clearer object contours without activating on background clutter and shows superior foreground-background separation. VajraV1-Xlarge also shows a more stable and smooth activation distribution compared to the uneven importance focus of the YOLOv12X \cite{tian2025yolov12attentioncentricrealtimeobject} and the scattered activations of the YOLOv11X \cite{yolo11_ultralytics}. It can also be observed from the first and last pictures that the VajraV1 exhibits higher semantic focus, consistently highlighting architecturally meaningful regions (such as domes, arches and windows) and has deeper vertical symmetry in activation while the YOLOv12X \cite{tian2025yolov12attentioncentricrealtimeobject} (while better than the YOLOv11X \cite{yolo11_ultralytics}) misses some structural lines. The improved performance of VajraV1-Xlarge can be attributed to its significantly larger receptive field.

\section{Conclusion}
It was observed that across multiple benchmarks, the \textbf{VajraV1-Nano, Small and Xlarge} consistently outperformed all of their YOLO counterparts while managing competitive latency figures. VajraV1-Medium outperformed the YOLOv12-M \cite{tian2025yolov12attentioncentricrealtimeobject} on the COCO Detection Benchmark \cite{lin2015coco}, performed at-par with the YOLOv12-M \cite{tian2025yolov12attentioncentricrealtimeobject} on the COCO Segmentation Benchmark \cite{lin2015coco}. While the VajraV1-Large performed slightly worse than the YOLOv12-L \cite{tian2025yolov12attentioncentricrealtimeobject} on both the COCO \cite{lin2015coco} Detection and Segmentation datasets. This is due to the larger receptive field of the YOLOv12-L when compared to the VajraV1-Large. However, on the Pose Estimation Benchmark, the VajraV1-Large achieved the same performance as the YOLOv11-X \cite{yolo11_ultralytics} while achieving comparable latency to the YOLOv11-L \cite{yolo11_ultralytics}.\\

VajraV1's improved performance over its YOLO counterparts is due to the various architectural improvements introduced in this paper. These are:

\begin{enumerate}
    \item Widening the 3x3 convolutions used in the primary computational block - VajraV1MerudandaX \cite{vajrav1_vayuai}
    \item Integrating a parameter efficient computational block by following YOLOv10's design philosophy of rank-guided design \cite{wang2024yolov10realtimeendtoendobject} - VajraV1MerudandaBhag15 \cite{vajrav1_vayuai}
    \item Utilizing FLOP and parameter efficient downsample convolutions - ADown \cite{wang2024yolov9learningwantlearn}
    \item Using a GELAN for integrating multiple Transformer modules - VajraV1AttentionBhag6 \cite{vajrav1_vayuai} (identical to C2PSA used in the YOLOv11 \cite{yolo11_ultralytics})
\end{enumerate}

\section{More Details}

VajraV1 was trained on the COCO Detection and Segmentation datasets for 600 epochs and on the COCO Pose dataset for 1000 epochs using the SGD optimizer with an initial learning rate of 0.01, momentum and weight decay set to 0.937 and 0.0005 respectively. The learning rate decays to $1 \times 10^{-4}$. VajraV1 has been trained using the same augmentations as the YOLOv11 \cite{yolo11_ultralytics} and the YOLOv12 \cite{tian2025yolov12attentioncentricrealtimeobject}. These include Mixup \cite{zhang2018mixupempiricalriskminimization}, Copy-Paste \cite{ghiasi2021simplecopypastestrongdata}, Mosaic \cite{bochkovskiy2020yolov4optimalspeedaccuracy, wang2022yolov7trainablebagoffreebiessets}. VajraV1 also utilised the Albumentations library for data augmentations \cite{info11020125}.

%\printbibliography
\bibliography{refs}  % Replace with your .bib file name

\end{document}